\documentclass[10pt,twocolumn,letterpaper]{article}

\usepackage{cvpr}
\usepackage{times}
\usepackage{epsfig}
\usepackage{graphicx}
\usepackage{amsmath}
\usepackage{amssymb}
\usepackage{booktabs}
\usepackage{lipsum,verbatim}

\usepackage[pagebackref=true,breaklinks=true,letterpaper=true,colorlinks,bookmarks=false]{hyperref}

\cvprfinalcopy 

\def\cvprPaperID{733}

\ifcvprfinal\fi
\begin{document}

\title{From {\em Lifestyle Vlogs} to Everyday Interactions}

\author{David F. Fouhey, Wei-cheng Kuo, Alexei A. Efros, Jitendra Malik\\
EECS Department, UC Berkeley}

\maketitle

\begin{abstract}
A major stumbling block to progress in understanding basic
human interactions, such as getting out of bed or opening a 
refrigerator, is lack of good training data. Most past efforts have 
gathered this data explicitly: starting with a laundry list of action labels, 
and then querying search engines for videos tagged with each label. In this 
work, we do the reverse and search implicitly: we start with a large collection of interaction-rich 
video data and then annotate and analyze it.  We use Internet Lifestyle Vlogs as the source of 
surprisingly large and diverse interaction data. 
We show that by collecting the data first, we are able to achieve greater scale 
and far greater diversity
in terms of actions and actors. Additionally,
our data exposes biases built into common explicitly gathered data.
We make sense of our data by analyzing the central component
of interaction -- hands. We benchmark two tasks: identifying
semantic object contact at the video level and non-semantic
contact state at the frame level. We additionally demonstrate
future prediction of hands.

\end{abstract}

\section{Introduction}
The lack of large amounts of good training data has long been a 
bottleneck for understanding basic everyday interactions. Past attempts to find this data have been largely unsuccessful:
there are large action recognition datasets but not for everyday interaction
\cite{UCF101,Kay17}, and laboriously obtained datasets
\cite{Sigurdsson2016,Koppula2013,Wu15,Goyal17} which depict everyday interaction,
but in which people are hired to act out each datapoint.

\begin{figure}
\includegraphics[width=\linewidth]{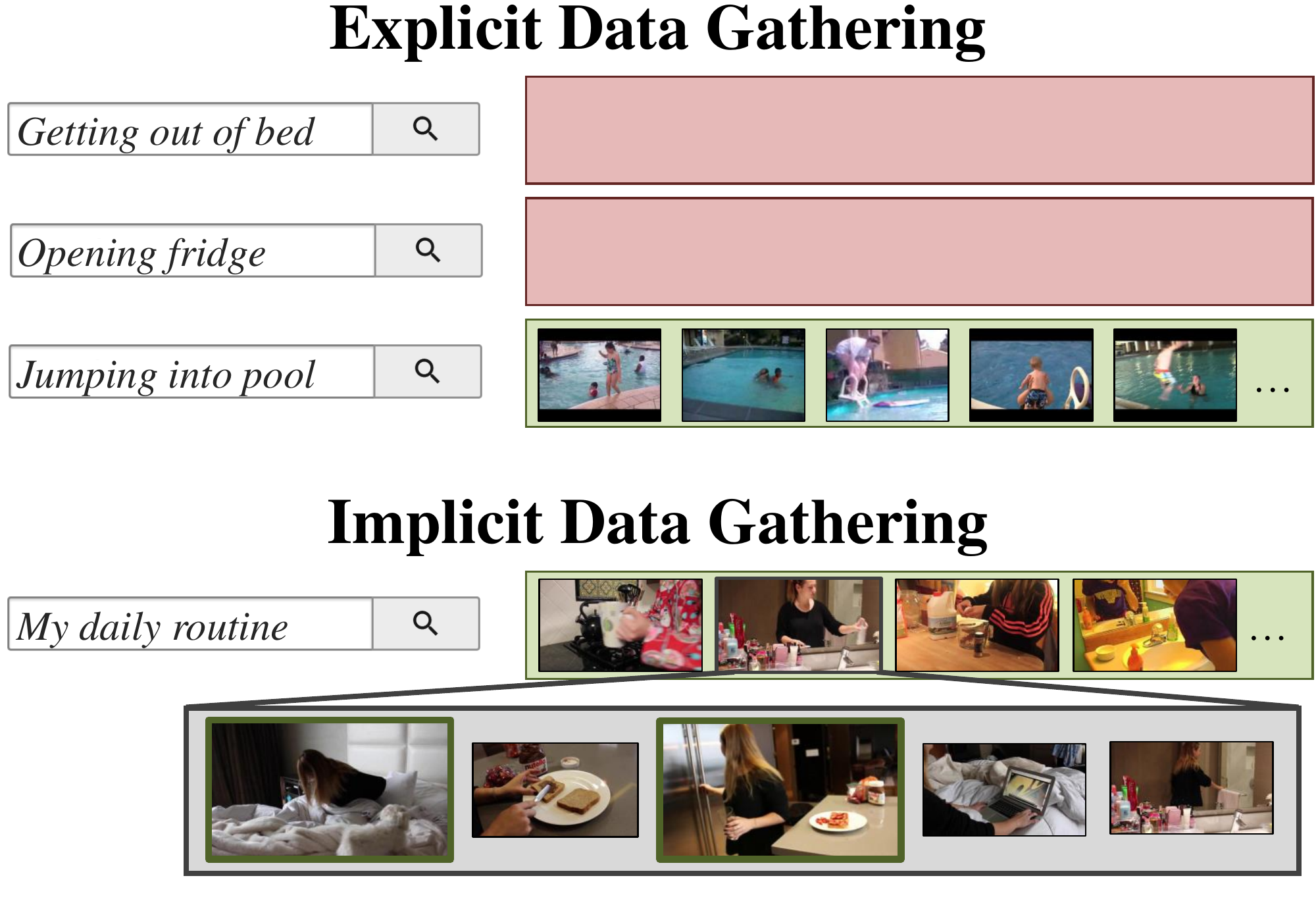}
\caption{Past work aimed at gathering everyday interaction data has been {\it explicit}, directly searching for a predetermined
list of categories. Unfortunately, direct search does not work for
everyday interactions like getting out of bed or opening a refrigerator since they are rarely tagged. Effort has thus
focused on things which are tagged, often unusual events. We present
{\it implicit} gathering as an alternative: everyday interactions exist buried in other data; we
can search for data that contains them and mine them. We demonstrate this by finding a new 14-day/114K video/10.7K uploader 
dataset of everyday interaction occurring naturally.}
\label{fig:teaser}
\end{figure}

\begin{figure*}
\includegraphics[width=\linewidth]{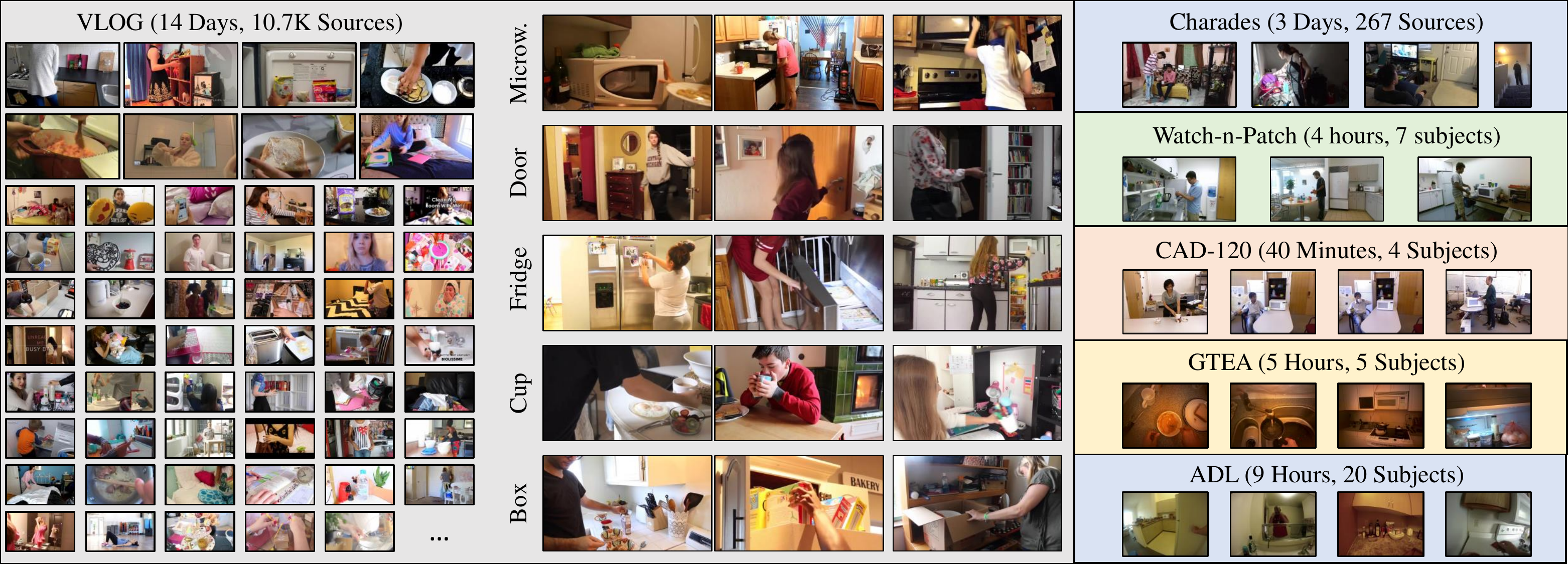}
\caption{{\bf An overview of our dataset VLOG,} which we
obtain by mining the vast amounts
of everyday interaction that exists implicitly in other data. We compare sample frames from
our dataset in comparison with video-collection-by-acting efforts such
as \cite{Sigurdsson2016,Koppula2013,Fathi2012,Pirsiavash12}.}
\label{fig:fig2}
\end{figure*}

The problem is that past methods have taken the approach of {\bf explicit} data gathering -- starting with a pre-determined taxonomy, they attempt to
directly find examples of each category. Along the way, they have fallen victim to 
dataset bias \cite{Torralba2011} 
in the form of a discrepancy between the world of reality and the {\it world of tagged things}.
This discrepancy dooms attempts to explicitly search for examples of everyday
interactions (``opening a microwave'', for instance, yields few good results 
\href{https://www.youtube.com/results?search_query=opening+a+microwave}{Try it!})
because there are few reasons to tag these videos.
Accordingly, most video efforts have focused on actions 
that can be found directly in the world of tagged things (e.g., high jump) 
as opposed to everyday ones that are impossible to find directly (e.g., opening a fridge).
Some researchers have identified this problem, and have proposed the solution
of {\it collection-by-acting} \cite{Sigurdsson2016,Koppula2013,Wu15,Goyal17}
in which people are hired to act out a script. This moves us considerably closer
towards understanding everyday interactions, but collection-by-acting is 
difficult to scale and make diverse.
But even if we ignore the struggle to find data, this explicit approach is
still left with two serious
bias issues, both of which we document empirically. First, the examples we recover
from the world of tagged things tend to be atypical: Internet search results
for a concept as basic as ``bedroom''
(\href{https://www.google.com/search?tbm=isch&q=bedroom}{Try it!}) are
hopelessly staged, taken at a particular distance, and almost always depict made beds.
Second, there are glaring gaps in terms of both missing categories and missing
negatives.

This paper proposes the alternative of {\it implicit} data gathering.
The intuition is that while we cannot find {\it tagged} everyday
interaction data, it exists implicitly inside other content 
that has different tags. We can find this superset of data, 
mine it for the data we want, and then annotate and analyze it. 
For us, this superset is {\it Lifestyle Vlogs}, videos that people 
purportedly record to show their lives.
As described in Section \ref{sec:collecting}, we mine this data
semi-automatically for interaction examples to produce {\bf VLOG},
a new large-scale dataset documenting everyday interactions. 
VLOG, illustrated in Figure \ref{fig:fig2}, is far larger and orders of magnitude more diverse than past
efforts, as shown in Section \ref{sec:dataset}. This shows
the paradoxical result that while implicit searching is less direct,
it is more effective at obtaining interaction data. 

While implicit gathering is more effective than explicit
gathering for interaction data, it also poses challenges. Explicitly gathered
data has a list of categories that predates the data, but implicitly gathered
data naturally depicts a long-tail of interaction types and must
be annotated post-hoc. We focus on the central figure of 
interaction, human hands and propose two concrete tasks in 
Section \ref{sec:labels}: (1) identifying 
contact state of hands in a video frame irrespective of object category,
which naturally covers the entire dataset; (2)
identifying if one of a number of objects was interacted with in the video.
This quantifies interaction richness, provides an index for other researchers,
and permits benchmarking of standard methods. We additionally provide labels
like scene categories and hand bounding-boxes that we use to explore our
data.

Our data and labels let us explore a large world of humans interacting
with their environment naturally. We first show that VLOG
reveals biases built into popular explicitly-gathered datasets in Section
\ref{sec:bias}. Having demonstrated this, we analyze how well current algorithms
work on our VLOG data and tasks in Section \ref{sec:baselines}. Finally, looking
towards the grand goal of understanding human interaction, we show applications of 
our data and labels for tasks like hand future prediction
in Section \ref{sec:exploring}.

\section{Related Work}

\begin{figure*}
\includegraphics[width=\linewidth]{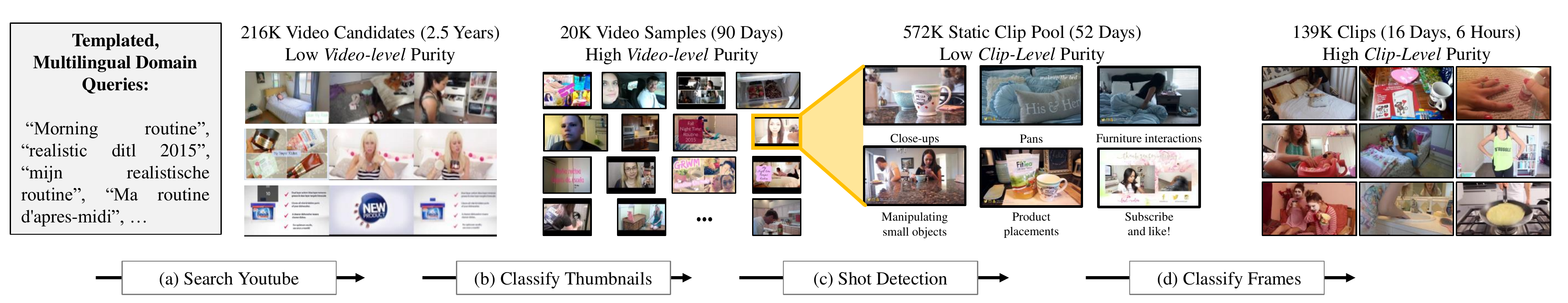}
\caption{ {\bf An illustration of the automatic gathering process.} Starting with a set
of templated queries, we (a) search YouTube (b) identify promising videos using
their thumbnails (c) break these videos into clips (d) identify promising 
clips. Throughout the process, the amount of data steadily decreases but
the purity steadily increases. Finally, the remaining clips are cleaned by humans.}
\label{fig:process}
\end{figure*}

This paper takes a step towards understanding everyday interactions and
thus touches on a number of areas of computer vision.

At the motivational level, the work is about affordances, or opportunities
for interaction with the world \cite{Gibson79}. This area has been extensively
studied in vision, typically aiming to infer affordances in still images
\cite{Grabner11,Gupta11,Roy2016,Yao10a,Yao2010}, understand scenes by observing them
over time \cite{Fouhey12,Delaitre12}, or use them as a building block for
scene understanding \cite{JiangSaxena13}. A fundamental stumbling block for
these efforts has been the difficulty of gathering interaction-rich video data 
at scale. While egocentric/life-logging efforts like 
\cite{Singh16,Rhinehart2016,Fathi2012,Pirsiavash12} do offer ways to obtain large
amounts of data in terms of volume, achieving diversity and coverage is an open challenge.
One contribution of this paper towards these efforts is demonstrating how to obtain
large scale interaction-rich data at scale while achieving diversity as well as a
concrete dataset that can be used to study humans ``in-the-wild'' from a variety of angles,
including benchmarks.

In this paper, we gather a large collection of videos  and annotate it
post-hoc with a variety of interaction labels. From this angle, our paper could
be viewed as similar to other action recognition efforts such as
\cite{Sigurdsson2016,Heilbron2015,Fathi2012,Rohrbach2012,Wang2016}.  The
primary distinction from this body of work is that we focus on everyday actions
and gather our data implicitly, rather than explicitly searching for it. Most
work focuses on non-everyday actions like ``high jump'' and work
focuses on everyday actions \cite{Sigurdsson2016,Koppula2013,Goyal17} gathers it explicitly by
acting. As we show, we can gather this data without searching for it directly,
achieving greater scale and diversity and, as we will show, avoiding some
sources of data bias.

\section{Collecting Interaction-Rich Episodes}
\label{sec:collecting}

We aim to get find data that is rich in everyday
interactions. As argued in the introduction, direct search
does not work, leading to efforts aimed at ``acting out'' daily activities 
\cite{Sung2012,Koppula2013,Sigurdsson2016,Goyal17}. By virtue
of their gathering approach, these datasets often have many desirable properties for
studying interaction compared to random Youtube videos. For instance,
many depict a single scene and feature a static camera, which make 
many learning tasks easier. We now show how to achieve a large scale while 
retaining features of
manual gathering efforts.

Our insight is that one can find interaction-rich genres 
and mine these videos for data. As a concrete example, we show
how to do this with a genre of YouTube video referred to as {\it
Lifestyle Vlogs} (Video-log\footnote{\noindent An archetypal example that
appears in our dataset is \url{https://youtu.be/DMZ_pRBd0dg}}). These are an immensely popular and multicultural
phenomenon of videos that purport to document the ordinary daily lives of their
recorders and feature titles like ``Daily Cleaning Routine!'' or ``A day in the
life of a stay at home mom''. As \cite{Kiberd2015} notes, the routines are, at
some level, aspirational as opposed to accurate. Nonetheless, even if the
series of actions (waking at 7AM with a cup of black coffee) represents an
ideal, at an interaction level (e.g., pouring a cup of coffee), the data is
realistic. Unfortunately, examples of interaction are interspersed
between camera pans of a well-kept house and monologues.

We thus distill our dataset semi-automatically, illustrated in Fig.~\ref{fig:process}. 
Each step hones in on an increasingly pure subset. We (i) we generate a large candidate video pool; 
(ii) filter with relevance feedback; (iii) 
segment videos into clips; (iv) filter clips by motion
and content; (v) filter the clips with annotators. Additional details
are in the supplemental.

\begin{table*}[t]
\caption[caption]{{\bf How VLOG compares to the most similar recent existing video datasets that
could be used to study everyday interactions. For comparison, we list representative
{\it action recognition} datasets with any overlap.} 
VLOG achieves the scale of many contemporary action recognition efforts
while also having features desirable for studying interaction: high demonstrator diversity, 
high resolution, static cameras, and open-world-like data. {\it Legend:} 
Diversity: \# of unique uploaders or actors; \%VGA+: what \% is
at least VGA resolution; 1st/3rd: person perspective; Implicit: whether the data
was gathered by explicitly finding the actions of interest.
}
\label{tab:dataCompare}
\centering
{\small 
\begin{tabular}{lc@{~~}c@{~~}ccc@{~~}cc@{~~}c@{~~}c} \toprule
            & \multicolumn{3}{c}{\bf Video Scale} & \bf Diversity 
            & \multicolumn{2}{c}{\bf Resolution} & \multicolumn{3}{c}{\bf Attributes} \\
Dataset     & Frames & Length & Count & Participants 
            & Mean & \% VGA+
            & 1st/3rd & Static & Implicit 
            \\
\midrule
VLOG & 37.2M & 14d, 8h & 114K & 10.7K & $660 \times 1183$ & 86\% & 3rd & $\checkmark$ & $\checkmark$ \\
\midrule
Something-Something \cite{Goyal17} & 13.1M & 5d, 1h & 108K & 1.1K & $100 \times 157$ & 0\% & 3rd & $\times$ & $\times$ \\
Charades \cite{Sigurdsson2016} & 8.6M & 3d, 8h & 9.8K & 267 & $671 \times 857$ & 56\% & 3rd & $\times$ & $\times$ \\
AVA \cite{Gu2017} & 5.2M & 2d & 192 & 192 & $451 \times 808$ & 26\% & 3rd & $\times$ & $\checkmark$ \\
Instructions \cite{Alayrac2016} & 795.6K & 7h, 22m & 150 & $< 150$ & $521 \times 877$ & 51\% & 3rd & $\times$ & $\times$ \\
Watch-n-Patch \cite{Wu15} & $78$K & 3h, 50m & 458 & 7 & $1920 \times 1080$ & $100\%$ & 3rd & $\checkmark$ & $\times$ \\
CAD-120 \cite{Koppula2013} & 61.5K & 41m & 120 & 4 & $480 \times 640$ & 100\% & 3rd & $\checkmark$ & $\times$ \\
GTEA \cite{Fathi2012} & 544.1K & 5h, 2m & 30 & 5 & $960 \times 1280$ & 100\% & 1st & $\times$ & $\checkmark$ \\
ADL \cite{Pirsiavash12} & 978.6K & 9h, 4m  & 20 & 20 & $960 \times 1280$ & 100\% & 1st & $\times$ & $\checkmark$ \\
MPI Cooking \cite{Rohrbach2012} & 881.8K & 9h, 48m & 5.6K &  12 & $1224 \times 1624$ & 100\% & 3rd & $\checkmark$ & $\checkmark$ \\

\bottomrule
\multicolumn{10}{c}{
{\bf $\uparrow$ Everyday Interaction $\uparrow$ ~~~~~~~~~~    $\downarrow$ Activity Recognition $\downarrow$ }} \\
\bottomrule
Kinetics \cite{Kay17} & 91M & 35d, 7h & 305K &  - & $658 \times 1022$ & $69\%$ & 3rd & $\times$ & $\times$ \\ 
ActivityNet \cite{Heilbron2015} & 69M & 27d, 0h & 20K & - & $640 \times 1040$ & $76\%$ & 3rd & $\times$ & $\times$ \\
UCF 101 \cite{UCF101} & 2.2M & 1d, 1h & 23K & - & $240 \times 320$ & $0\%$ & 3rd & $\times$ & $\times$ \\
\bottomrule
\end{tabular}
}
\vspace{-0.15in}
\end{table*}

\noindent {\bf Finding Videos.}
We first find a Lifestyle Vlog corpus. We define a positive video as
one that depicts people interacting with the indoor environment from a
3rd person. We additionally exclude videos
only about makeup and ``unboxing'' videos about purchases.

We use templated queries based on themes (``daily routine 2013'') or activities
involved (``tidying bedroom''), including 6 main English query templates and 3
templates translated into 13 European languages. These give 823 unique queries.  We
retrieve the top 1K hits on YouTube, yielding 216K unique candidate videos.
The results are 23\% pure at a video level: failures include
polysemy (e.g., ``gymnastic routine''), people talking about 
routines, and product videos. 

This candidate corpus is too large ($\sim$14TB) and noisy to contemplate downloading
and we thus filter with the four thumbnails that can be
fetched from YouTube independently of the video.  We labeled 1.5K videos as
relevant/irrelevant. We then represent each video by summary statistics of its
ILSVRC-pretrained \cite{ILSVRC15} Alexnet \cite{Krizhevsky12} {\tt pool5}
activations, and train a linear SVM. We threshold and retrieve 20K videos.

\noindent {\bf Finding Episodes Within Videos.}
This gives an initial dataset of lifestyle Vlogs with high purity at 
the {\it video level}; however, the interaction events 
are buried among a a mix of irrelevant sequences and camera motion.

We first segment the videos into clips based on camera motion. Since
we want to tag the start of a slow pan while also not cutting at
a dramatic appearance change due to a fridge opening, we use 
an approach based on homography fitting on SIFT \cite{Lowe2004} matches. 
The homography and SIFT match count are used to do shot as well as static-vs-moving
detection. After discarding clips shorter than 2s, this yields a set of 572K
static clips.

Many remaining clips are irrelevant, such as 
subscription pleas or people talking to the
camera. Since the irrelevant clips resemble the irrelevant videos, we
reuse the SVM on CNN activations approach for filtering.
This yields 139K clips that mostly depict human interactions.

\noindent {\bf Manual Labeling.} 
Finally, workers flag 
adult content and videos without humans touching something with their hands.
This yields our final set of 114K video clips, showing the automatic stages
work at $82\%$ precision.

\section{The VLOG Dataset}
\label{sec:dataset}

We now analyze the resulting underlying data in context of past efforts. 
{\bf We will freely release this data along with all annotations
and split information to the community.}
Here, we focus on the data itself: as a starting point, we
provide a number of annotations; however, since our data was gathered 
before our labels were defined, the data
can be easily relabeled, and the videos themselves can serve as labels
for unsupervised learning.

As shown in Table \ref{tab:dataCompare}, VLOG is closer in sheer volume to
traditional activity datasets like ActivityNet and two days longer than
all of the every other dataset listed.
However, VLOG is distinguished not just in size but also in 
diversity: it has 
$\approx9.4\times$ more source than Something-Something and
$\approx40\times$ more sources than Charades (and is more balanced
in terms of uploaders, Gini coefficient $0.74$ vs $0.57$). 
We can put this diversity in perspective by calculating how many frames
you could sample before expecting to see the same person (for datasets
where this information is available). In CAD-120, it is
just $2$; Watch-n-Patch is $3$; Charades is $10$; and VLOG $58$. We 
report additional dataset statistics about VLOG (e.g., scene types, distribution of 
video length) in the supplemental.

Compared to direct gathering efforts (e.g., CAD-120) in which
there are direct incentives for quality, crawling efforts come at the cost
of a lack of control. Nonetheless, our average resolution approaches
that of in-lab efforts. This is because our content creators are
motivated: some intrinsically  and some because they make money via advertising. Indeed, many videos
are clearly shot from tripods and, as the figures throughout the paper show,
most are lit and exposed well. 

Our paper is best put in context with video datasets, but of course 
there are image-based interaction datasets such as HICO
\cite{Chao15} and V-COCO \cite{Gupta15}. 
As image-based data, though, both depend on someone taking, uploading, and
tagging a photo of the interaction. Accordingly,
{\it despite directly searching} for
refrigerator and microwave interactions, HICO contains only 59 and 22
instances of each. VLOG, as we will next see, has far more despite
directly searching for neither.

\section{Labels}
\label{sec:labels}

While implicitly gathered data scales better, it presents challenges for
annotation. Explicitly gathered data naturally maps to categories and tasks
since it was obtained by finding data for these categories; implicitly
gathered data, on the other hand, naturally depicts a long tail distribution.
We can quantify this in VLOG by analyzing the entry-level categories (examples appear in 
supplemental) being interacted with
in a $500$ image sample: standard species richness estimation techniques 
\cite{Burnham78} give an estimate of $346$ categories in the dataset. 
This is before even distinguishing objects more finely (e.g., wine-vs-shampoo
bottles) or before distinguishing interactions in terms of verbs (e.g.,
open/pick up/pour from bottle).

Faced with this vast tail, we focus on the crucial part of the interaction,
the hands, and pose two tasks. The first is whether the hands interact
with one of a set of semantic objects in the clip. 
As a side benefit, this helps quantify our data in comparison to 
explicit efforts and gives an index into the dataset, which is useful 
for things like imitation learning. The second task is the contact state 
of the hands at a frame level. This describes human behavior in a way
that is agnostic to categories and therefore works across all object categories in the dataset.

In addition to labels, we use the YouTube uploader id to define 
standard $50/25/25$ train/val/test splits where
each uploader appears only in one split.

\subsection{Hand/Semantic Object Annotations}

\begin{figure}[t]
\includegraphics[width=\linewidth]{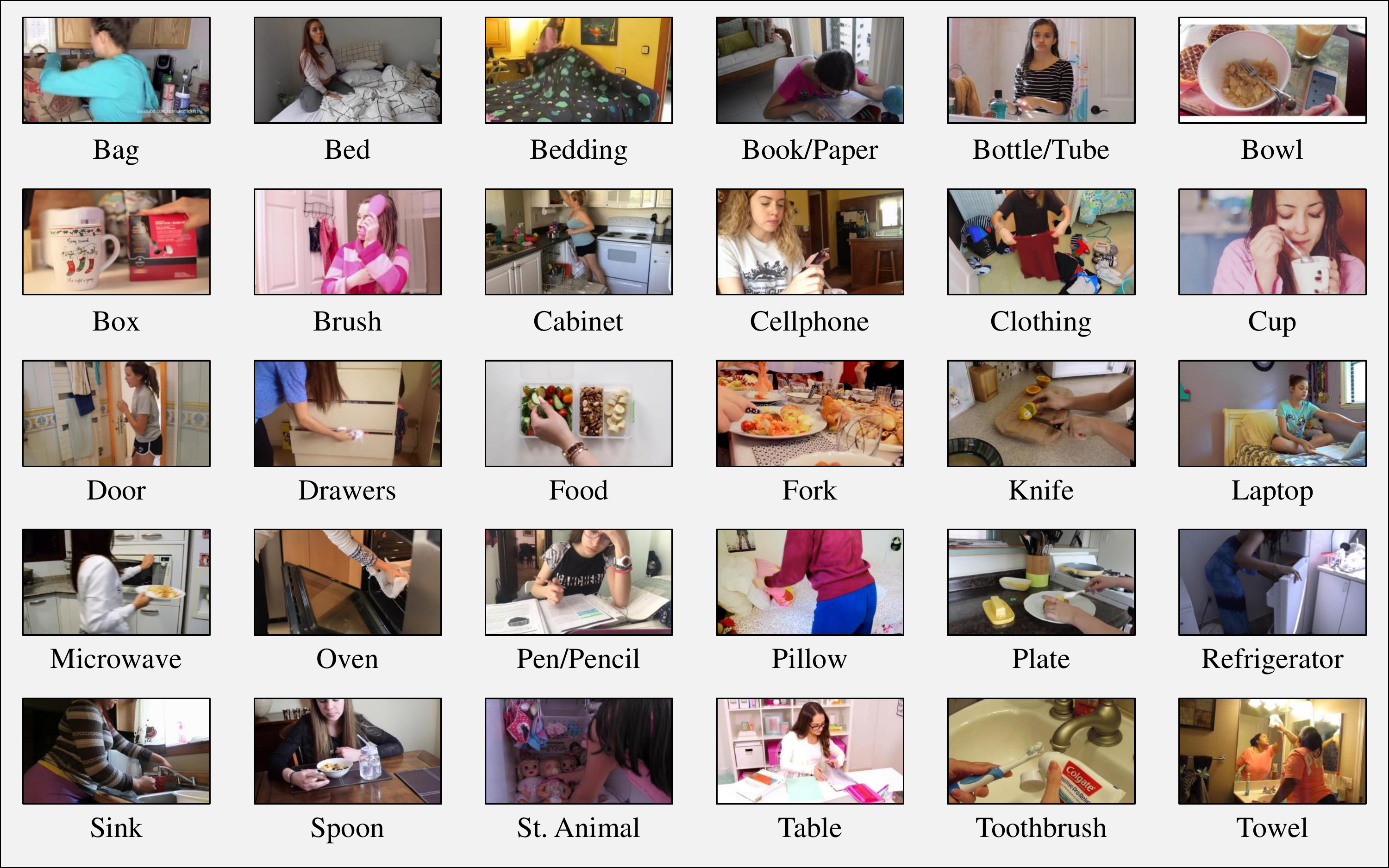}
\caption{Examples of each of our 30 objects being interacted with.}
\label{fig:annot_sampler}
\vspace{-0.1in}
\end{figure}

We frame this as whether a human interacts with any instance of each of a set
of $30$ objects (i.e., {\it $30$ binary, clip-level tasks}).  We focus on
clip-level annotation because our many of our clips are short, meaning
that clip-level supervision is quite direct in these cases. Note 
independent tasks are necessary since people may do multiple things.

\noindent {\bf Vocabulary.} 
Since our data was collected implicitly,
we must determine a vocabulary.
An entirely new one would preclude studying transfer, but 
an entirely existing one would spend effort on objects that are not
present. We resolve these competing aims by taking an empirical approach but
favoring COCO objects. We exhaustively described which objects were interacted
with in a subset of our data; we select $30$ categories by identifying frequent
objects specific to our dataset (e.g., bedding, doors) and COCO objects that
are sufficiently frequent (e.g., fridges, microwaves). 

\noindent {\bf Annotation.} We asked workers to annotate the videos
at a clip level based on whether the human made hand contact with an instance
of that object category. Following \cite{sigurdsson2016much}, multiple
annotators were asked to annotate a few objects (8) after watching a video. 
Video/object set pairs where annotators could not come to a consensus were marked
as inconclusive. This and all labeling was done through a crowdsourcing service, which
used standard quality-control such as consensus labeling, qualifications, and
sentinels. Sample labels are shown in Fig.~\ref{fig:annot_sampler}

\begin{figure}[t]
\includegraphics[width=\linewidth]{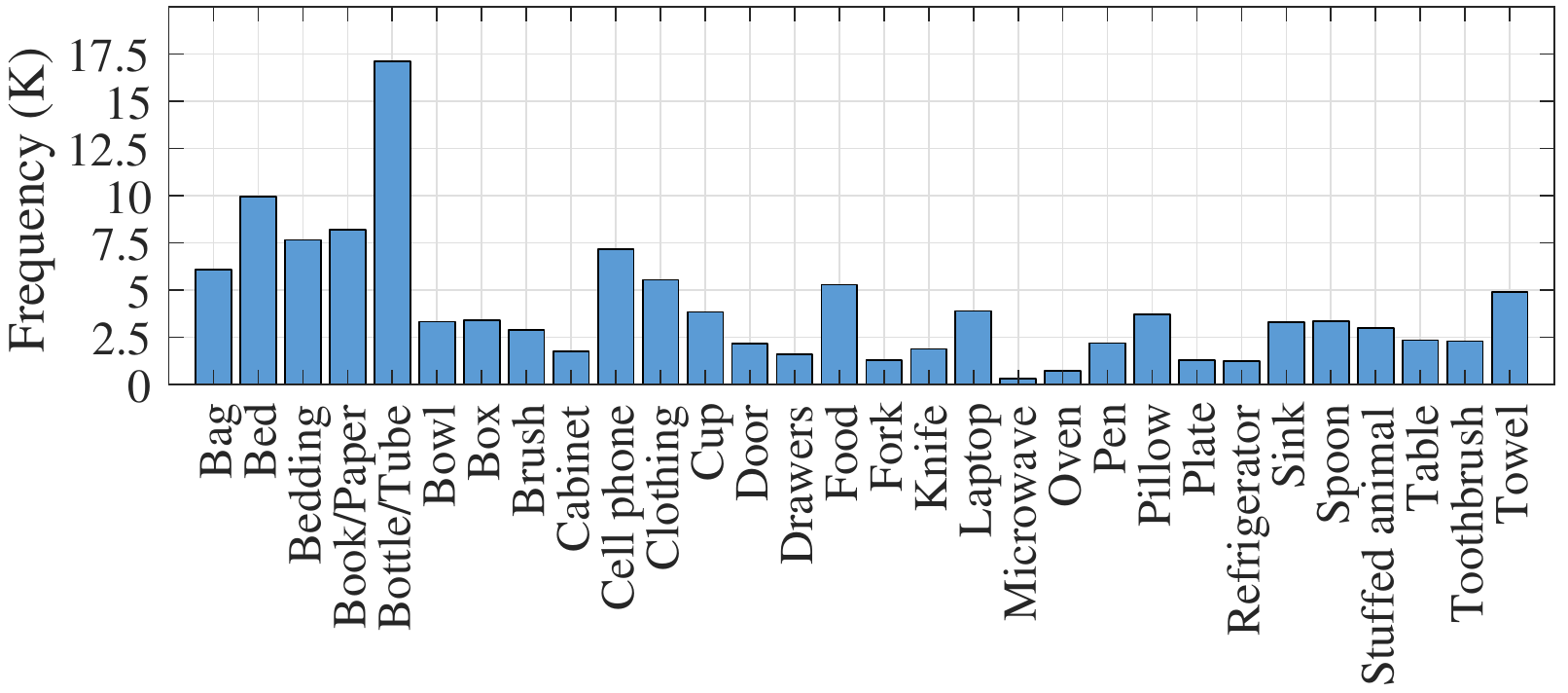}
\vspace{-0.1in}
\caption{Hand/Semantic Object Label frequency}
\label{fig:semantic_label}
\vspace{-0.2in}
\end{figure}

\noindent {\bf Labels.} 
Fig.~\ref{fig:semantic_label} shows that the human/object interactions are unevenly distributed.
Microwaves, for instance, are interacted with far less
frequently ($<0.3\%$) than cell phones or beds.  Nonetheless, 
there are $296$ instances, making it the largest collection 
available. Moreover, since this was obtained {\it without 
searching for} microwave, we expect similar quantities of many
other objects can be obtained easily.

\noindent {\bf Comparison.} 
We now compare the scope of VLOG in terms of object interaction 
with any existing dataset. We examined 15 VLOG categories that overlap
cleanly with any of \cite{Sigurdsson2016,Koppula2013,Wu15,Chao15,Gupta15}.
First, most datasets have catastrophic gaps:
\cite{Sigurdsson2016} has $2$ microwaves, for instance, and \cite{Koppula2013} has no laptops
among many other things.
Averaging across the categories, VLOG has 
$5\times$ the number of examples compared to the largest of past work. 
This measure, moreover, does not account for diversity: all $36$ microwave videos of
\cite{Koppula2013} depict the same exact instance, for example. Bed is
the largest relative difference since many source videos start in the morning.
The only category in which VLOG lags is doors with only $2179$ compared to Charades'
$2663$. 

\subsection{Hand Contact State Annotations}

We then annotate the hand contact state of a large set of frames. 
This automatically entirely covers the space of videos:
irrespective of which object is being interacted with, we can identify that there
is interaction. 

\noindent {\bf Vocabulary.} Our vocabulary first identifies whether
0, 1, or $2+$ people are present; images with $1$ person
are labeled with how many hands are {\it visible} and how many (but not which) visible
hands are in contact, defined as touching something other than the human's
body or worn clothing. This gives $6$ hand states and $8$ total categories.

\noindent {\bf Annotation.} We annotate a random $219K$ subset of frames;
images without worker agreement are marked as inconclusive.  We can trivially
convert these labels into ``future'' labels for problems like contact
anticipation.

\subsection{Additional Data Annotations}

Finally, to better understand the nature of the data, we additionally annotated
a ``taster'' 20K subset of it with additional labels. We use these
to better understand performance. Since our
videos are single scenes taken from a static camera, we mark
keyframes from the middle of the video. The one exception is
object presence, where we annotate videos to match
the style of our primary annotations.

\noindent {\bf Object Presence.} We additionally annotate each of whether $30$ classes
{\it appears} in the video to distinguish contact from presence.  Since
annotators search rather than watch the hands, 
recall drops; however, precision remains high.

\noindent {\bf Scene Classification.} We annotate scenes as being
shot in one of $5$ categories (bathroom, bedroom, dining room, kitchen, 
living room) or none-of-the-above. The five categories cover $76\%$ of our
data.

\noindent {\bf Proxemics.} We annotate distance to the
nearest object via Hall's proxemic categories \cite{Hall66}:
intimate ($<.45\textrm{m}$), personal ($<1.2\textrm{m}$), social ($<3.7\textrm{m}$), 
public ($3.7\textrm{m}+$).

\noindent {\bf Human Configurations.} We annotate the visibility of
the head, torso, hands, and feet by categorizing the image into the six common
visibility configurations (capturing $92\%$ of the data), as well as none of
the above or no human visible.

\noindent {\bf Hand Location.} We further annotate 5K images with
bounding box locations of hands from our contact images.

\section{Exploring Biases of Explicit Data}
\label{sec:bias}

We first examine to what extent current recognition systems
can make sense of VLOG by applying standard models for 
scene classification and object detection that were
trained on standard datasets (both gathered explicitly). Before describing the
experiments, we note that VLOG has no blatant domain shift issues: it shows
objects and scenes in normal configurations shot from real sensors with little
blur. Nonetheless, our experiments show failures that we trace 
back to biases caused by explicit gathering. 

\begin{table*}
\caption{{\bf Results on Hand/Semantic Object Interaction Classification (Average precision).} }
\label{tab:humanobject}
\begin{tabular}{l@{  }c@{  }c@{  }c@{  }c@{  }c@{  }c@{  }c@{  }c@{  }c@{  }c@{  }c@{  }c@{  }c@{  }c@{  }c@{  }c@{  }c@{  }c@{  }c@{  }c@{  }c@{  }c@{  }c@{  }c@{  }c@{  }c@{  }c@{  }c@{  }c@{  }c@{  }c}
\toprule
  & \small \rotatebox{90}{mAP}  & \small \rotatebox{90}{bag} 
 & \small \rotatebox{90}{bed} 
 & \small \rotatebox{90}{bedding} 
 & \small \rotatebox{90}{book/papers} 
 & \small \rotatebox{90}{bottle/tube} 
 & \small \rotatebox{90}{bowl} 
 & \small \rotatebox{90}{box} 
 & \small \rotatebox{90}{brush} 
 & \small \rotatebox{90}{cabinet} 
 & \small \rotatebox{90}{cell-phone} 
 & \small \rotatebox{90}{clothing} 
 & \small \rotatebox{90}{cup} 
 & \small \rotatebox{90}{door} 
 & \small \rotatebox{90}{drawers} 
 & \small \rotatebox{90}{food} 
 & \small \rotatebox{90}{fork} 
 & \small \rotatebox{90}{knife} 
 & \small \rotatebox{90}{laptop} 
 & \small \rotatebox{90}{microwave} 
 & \small \rotatebox{90}{oven} 
 & \small \rotatebox{90}{pen/pencil} 
 & \small \rotatebox{90}{pillow} 
 & \small \rotatebox{90}{plate} 
 & \small \rotatebox{90}{refrigerator} 
 & \small \rotatebox{90}{sink} 
 & \small \rotatebox{90}{spoon} 
 & \small \rotatebox{90}{stuffed animal} 
 & \small \rotatebox{90}{table} 
 & \small \rotatebox{90}{toothbrush} 
 & \small \rotatebox{90}{towel} 
\\ 
\midrule
\small Key R50  & \scriptsize 31.4  & \scriptsize 21.7  & \scriptsize 62.5  & \scriptsize 57.6  & \scriptsize 51.2  & \scriptsize 51.0  & \scriptsize 25.7  & \scriptsize 17.3  & \scriptsize 11.2  & \scriptsize 16.4  & \scriptsize 40.0  & \scriptsize 34.0  & \scriptsize 19.6  & \scriptsize 34.8  & \scriptsize 19.8  & \scriptsize 40.3  & \scriptsize 12.5  & \scriptsize 26.1  & \scriptsize 48.1  & \scriptsize 39.3  & \scriptsize 20.6  & \scriptsize 24.1  & \scriptsize 33.2  & \scriptsize 7.2  & \scriptsize 46.4  & \scriptsize 49.4  & \scriptsize 24.4  & \scriptsize 45.2  & \scriptsize 19.9  & \scriptsize 24.5  & \scriptsize 17.9 \\ 
\small $\mu$ R50  & \scriptsize 36.2  & \scriptsize 27.1  & \scriptsize 67.4  & \scriptsize 63.0  & \scriptsize 58.0  & \scriptsize 56.0  & \scriptsize 29.4  & \scriptsize 20.3  & \scriptsize 17.4  & \scriptsize 20.1  & \scriptsize 46.1  & \scriptsize 39.5  & \scriptsize 21.8  & \scriptsize 45.0  & \scriptsize 25.3  & \scriptsize 45.5  & \scriptsize 14.8  & \scriptsize 30.2  & \scriptsize 54.5  & \scriptsize 42.5  & \scriptsize 22.2  & \scriptsize 30.1  & \scriptsize 36.4  & \scriptsize 8.7  & \scriptsize 52.7  & \scriptsize 54.6  & \scriptsize 28.0  & \scriptsize 52.4  & \scriptsize 21.2  & \scriptsize 33.3  & \scriptsize 23.6 \\ 
\small I3D K  & \scriptsize 28.1  & \scriptsize 20.2  & \scriptsize 56.9  & \scriptsize 54.3  & \scriptsize 47.5  & \scriptsize 45.3  & \scriptsize 21.8  & \scriptsize 14.8  & \scriptsize 22.8  & \scriptsize 16.3  & \scriptsize 36.6  & \scriptsize 32.5  & \scriptsize 16.6  & \scriptsize 32.7  & \scriptsize 12.5  & \scriptsize 36.2  & \scriptsize 12.6  & \scriptsize 29.4  & \scriptsize 36.0  & \scriptsize 17.7  & \scriptsize 13.2  & \scriptsize 25.7  & \scriptsize 30.5  & \scriptsize 6.6  & \scriptsize 27.3  & \scriptsize 45.5  & \scriptsize 23.9  & \scriptsize 30.1  & \scriptsize 15.1  & \scriptsize 39.1  & \scriptsize 24.2 \\ 
\midrule\small FT R50  & \scriptsize 40.5  & \scriptsize \bf 29.7  & \scriptsize 68.9  & \scriptsize 65.8  & \scriptsize \bf 64.5  & \scriptsize \bf 58.2  & \scriptsize \bf 33.1  & \scriptsize \bf 22.1  & \scriptsize 19.0  & \scriptsize \bf 23.9  & \scriptsize \bf 54.0  & \scriptsize 45.5  & \scriptsize \bf 28.6  & \scriptsize 49.2  & \scriptsize \bf 28.7  & \scriptsize 49.6  & \scriptsize \bf 19.4  & \scriptsize 37.5  & \scriptsize \bf 62.9  & \scriptsize \bf 48.8  & \scriptsize \bf 23.0  & \scriptsize 36.9  & \scriptsize 39.2  & \scriptsize 12.5  & \scriptsize \bf 55.9  & \scriptsize 58.8  & \scriptsize 31.1  & \scriptsize \bf 57.4  & \scriptsize \bf 26.8  & \scriptsize 39.6  & \scriptsize 22.9 \\ 
\small FT I3D-K  & \scriptsize 39.7  & \scriptsize 24.9  & \scriptsize \bf 71.7  & \scriptsize \bf 71.4  & \scriptsize 62.5  & \scriptsize 57.1  & \scriptsize 27.1  & \scriptsize 19.2  & \scriptsize \bf 33.9  & \scriptsize 20.7  & \scriptsize 50.6  & \scriptsize \bf 45.8  & \scriptsize 24.7  & \scriptsize \bf 54.7  & \scriptsize 19.1  & \scriptsize \bf 50.8  & \scriptsize 19.3  & \scriptsize \bf 41.9  & \scriptsize 54.0  & \scriptsize 27.5  & \scriptsize 21.4  & \scriptsize \bf 37.4  & \scriptsize \bf 42.9  & \scriptsize \bf 12.6  & \scriptsize 42.5  & \scriptsize \bf 60.4  & \scriptsize \bf 33.9  & \scriptsize 46.0  & \scriptsize 23.5  & \scriptsize \bf 59.6  & \scriptsize \bf 34.7 \\ 
\bottomrule 
\end{tabular}

\vspace{-0.1in}
\end{table*}

\noindent {\bf Scene Classification.} We take the public
Densenet-161 \cite{huang17} model trained on the 1.8M image 
Places365-Standard dataset \cite{zhou17} and apply it
to VLOG. Specifically we classify each
frame labeled with scene class into 365 scene categories from
Places365. We quantify performance with the top-5 accuracy. 

\begin{figure}
\includegraphics[width=\linewidth]{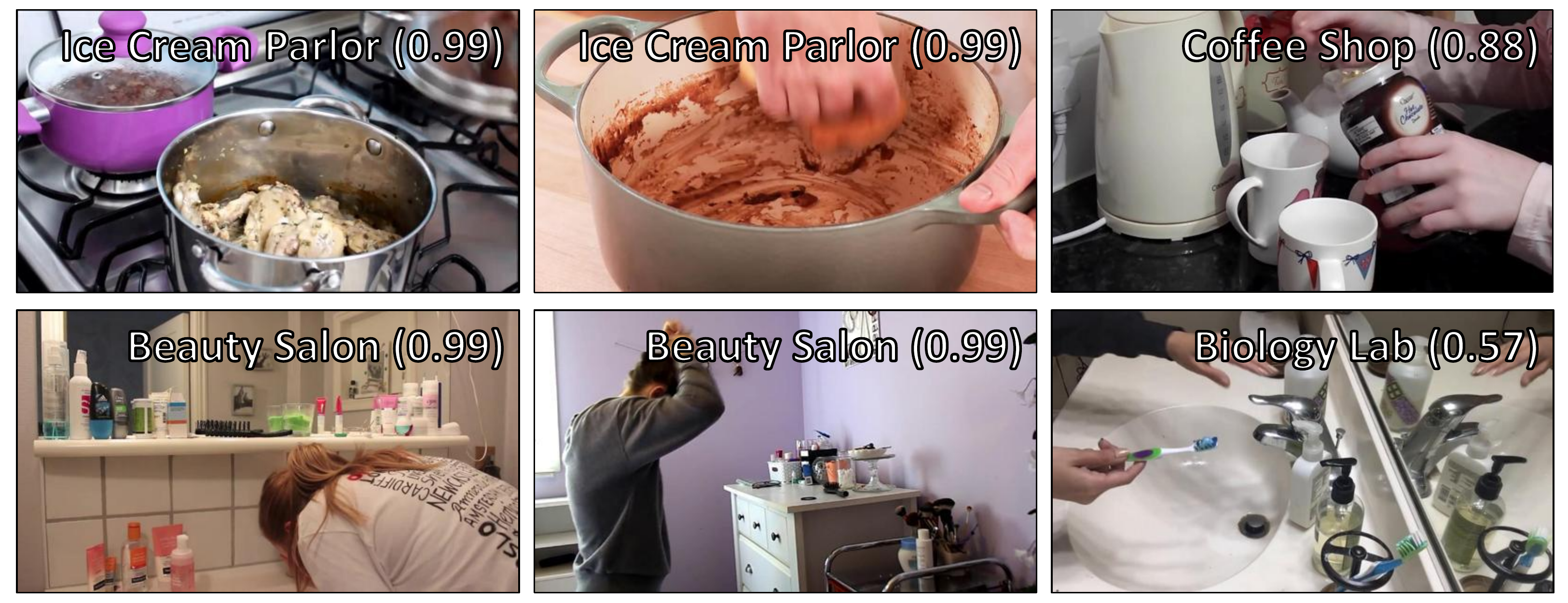}
\caption{We show examples of frequent confusions
as well as their top prediction and confidence. Often the model is confident
and wrong; sometimes it is baffled by easy images.}
\label{fig:scene_mistakes}
\vspace{-0.15in}
\end{figure}

\begin{figure}
\includegraphics[width=\linewidth]{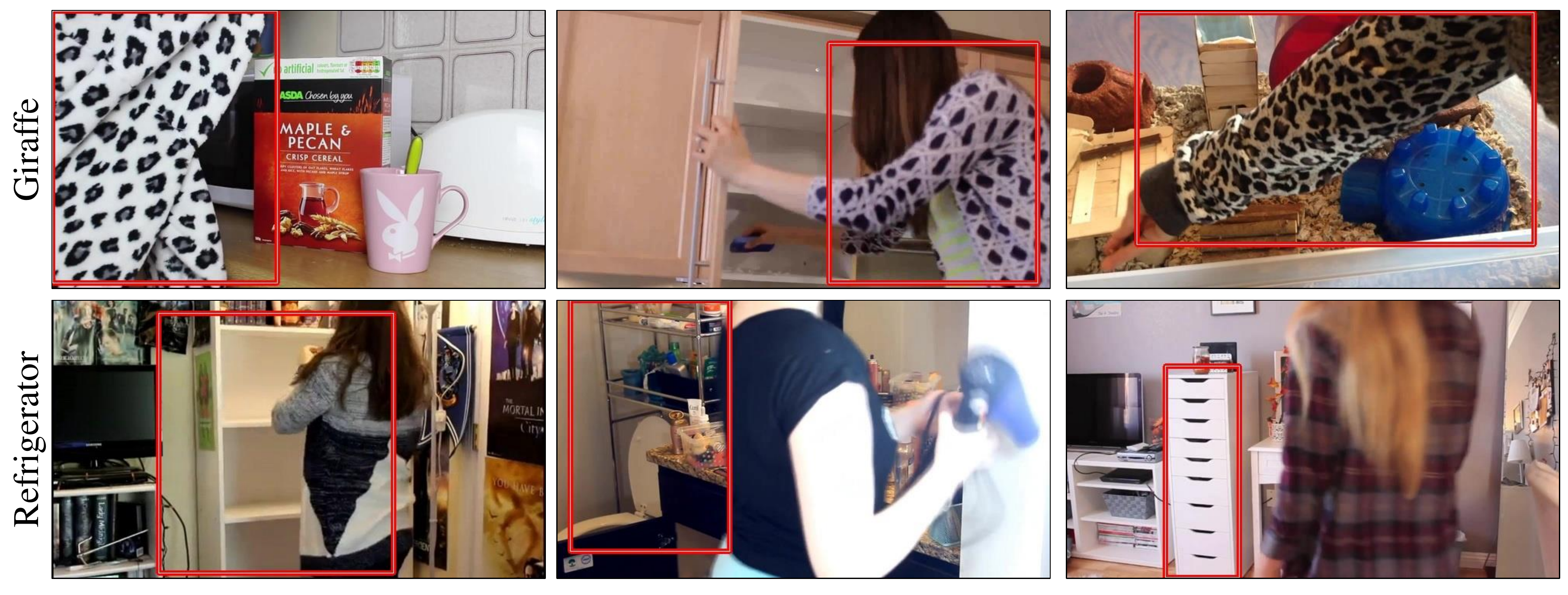}
\caption{{\bf Detection False Positives.} We show sample high-confidence 
(${>}0.9$) detections from Faster RCNN on VLOG. These suggest COCO contains
shortcut solutions to some classes, namely blobby textures for giraffes
and shelves for refrigerators.}
\label{fig:detectionMistakes}
\vspace{-0.15in}
\end{figure}

The off-the-shelf network struggles: in contrast
to $85\%$ top-5 accuracy on the original dataset,
it gets just $43\%$ on VLOG.  The degradation is not graceful: kitchens
are often seen as laboratories or an ice cream parlor
(sample mistakes are in Fig~\ref{fig:scene_mistakes}), and reaching
$80\%$ accuracy requires using the top-28 error.
One hypothesis might be that VLOG is just intrinsically harder: however,
humans were all able to come to a consensus on these frames, and 
a simple model with no fine-tuning (linear SVM on pretrained 
Resnet-50 \cite{He2015} activations) is able to achieve $70\%$ top-1 accuracy.

The cause is an enormous domain shift from the world of 
images {\it tagged} with scene classes to the world of images from those scenes.
Examining the source dataset, Places365, reveals sterile scenes of kitchens with uncluttered
counters and bedrooms with made beds, taken from a distance to show off the scene (samples in the
supplemental). It is no surprise then that the network fails on views of a
dresser in a bedroom or an in-use stovetop. We can verify this intuition by the steep dropoff in accuracy 
as one gets closer to the scene: at social/personal/intimate distance, accuracy is 
$66.5\%$/$48.2\%$/$28.2\%$.

\noindent {\bf Object detection.} We take the publicly available 
VGG-16 \cite{Simonyan14c}-based faster RCNN \cite{Ren2015} 
network. This was trained on COCO \cite{Lin2014} to detect 80 categories of objects.
We run this detector at $3$Hz and max-aggregate over the video. 

We find a number of failure modes that we trace back to a
lack of the right negatives.
Fig.~\ref{fig:detectionMistakes} shows sample
confident detections for giraffe and refrigerator; these are thresholded
at ${>}0.9$, corresponding to ${>}99\%$ and ${>}96\%$ precision on COCO and
come from a larger set of false positives on blobby textures and shelf-like
patterns. Since VLOG has many refrigerators, we can quantify performance at
this operating point for refrigerators. We count a detection as correct if
it contains the object of interest: the $96\%$ precision on COCO (computed the
same way) translates to far worse $44\%$ precision on VLOG, with similar recall. 

We hypothesize these failures occur because of missing negatives due to 
explicit gathering. COCO was gathered explicitly looking for giraffes rather
than documenting the savannah and so there are no leopards to force the network to
go beyond texture classification. We find similar false positive issues for
zebras, whose texture is distinctive, but not for bears (whose texture
must be distinguished from dogs). Similarly, most refrigerator false positives
are photos that are unlikely -- e.g., cleaning an empty bookshelf. Finally, we note
that finding this out via COCO is difficult -- giraffe has the highest overall
AP for the method, and refrigerator is the second highest of the appliances.

\section{Benchmarking Interactions}
\label{sec:baselines}

Now that we have analyzed some difficulties current off-the-shelf models 
have in interpreting VLOG, we analyze how well we can learn to understands
hands interacting with the world in VLOG. Our goal in this section is to 
understand how well {\it existing} techniques work on VLOG; introducing 
new architectures for video and image understanding is beyond the scope
of this work, although our error modes suggest likely future directions.

\subsection{Hand/Semantic Object Interaction}

\begin{figure*}
\centering
\includegraphics[width=\linewidth]{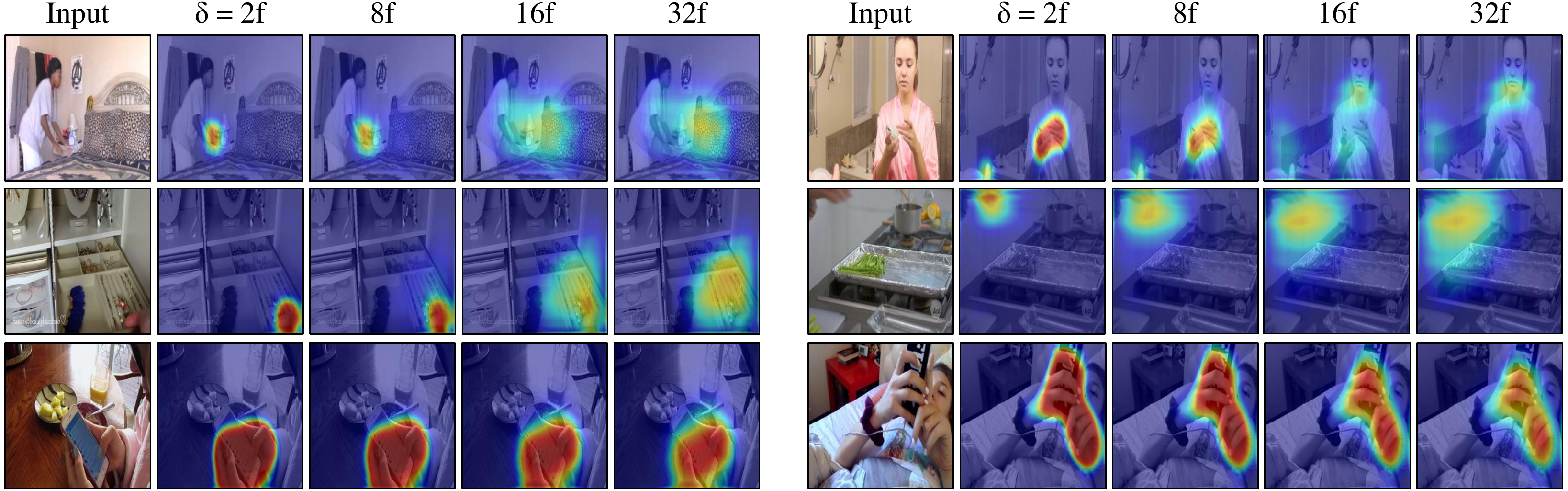} 
\caption{{\bf Where will the hands go?} By having an enormous dataset
of hands interacting with the world, we can begin learning models of how
everyday interactions play out. Here, we show outputs from a model that 
predicts where hands will be in the future $2$, $8$, $16$, and $32$ frames in the
future. It has learned plausible dynamics,
including identifying that people using cellphones are unlikely to put them down.}
\label{fig:predictions}
\vspace{-0.15in}
\end{figure*}

We first analyze our human-object contact benchmark, a set of 30 video-level
binary decisions. We quantify performance with average precision.

\noindent {\bf Models.} Our models are inspired by what was done in Charades \cite{Sigurdsson2016}.
We begin with single frame models.
The first two use a linear SVM on aggregated final-layer activations of a single-frame
ILSVRC-pretrained \cite{ILSVRC15} Resnet-50. We try the following, all L2-normalized: {\it (Key R50)} one middle frame, which
shows how much is explained by a scene glimpse; {\it ($\mu$ R50)}
the mean of the feature vector over time. 
Finally, we fine-tune the model  on
VLOG ({\it FT R50}); at test time, we average over evenly spaced locations.  
We next use standard action recognition, using the Kinetics-pretrained \cite{Kay17} RGB version of
I3D \cite{Carreira17}, the class of models that is state of the art on
\cite{Sigurdsson2016}. We train a linear SVM on average activations ({\it I3D-K});
as well as fine-tune it on VLOG ({\it FT I3D-K}). Note the base architecture,
Inception-v1, has lower spatial resolution and depth than Resnet50.

\noindent {\bf Results.} We show quantitative results in
Table~\ref{tab:humanobject}. Fine-tuning improves results, and
the I3D exhibits far larger gains compared to Resnet50,
suggesting a large domain gap between Kinetics and VLOG. 
Some objects, mainly textureless or small ones interacted with
in distinctive motions  -- bedding, brushes, toothbrushes, and towels --
benefit tremendously from the temporal signal in I3D. Others, usually appliances 
like microwaves and refrigerators, benefit from Resnet50's higher spatial resolution
and depth.

We see a number of areas for improvement. A single frame 
does poorly but also gets $78\%$ of the performance relative to the best method, suggesting 
that current techniques are missing lots of information. Further analysis
shows that mAP drops at social distance:
FT I3D-K $39.7\% \to 27.0\%$, with large drops for manipulated objects
like cups or knives. This suggests we may need architectures that can
transfer from up-close examples to far-away ones or focus
more closely on hands.

\subsection{Hand Contact State}

\begin{table}
\caption{Accuracy for hand-state prediction in the present as well as 
$6$, $12$, $30$, and $60$ frames in the future.}
\centering
\label{tab:handstate}
\begin{tabular}{lccccc} \toprule
                & \small Now & \small +6f & \small +12f & \small +30f & \small +60f \\ \midrule
\small R50             & \small 43.6 & \small 41.9 & \small 40.4 & \small 37.5 & \small 35.7
\\
\small FT R50          & \small 56.4 & \small 49.6 & \small 45.9 & \small 41.0 & \small 37.8
\\
\small FT R50+Pseudo Labels  & \small \bf 58.2 & \small \bf 53.1 & \small \bf 49.6 & \small \bf 43.8 & \small \bf 39.4
\\
\bottomrule
\end{tabular}
\vspace{-0.1in}
\end{table}

While semantic labels provide one view of our data, VLOG's long-tail ensures that
many interactions will go uncategorized since they do not fall into one of the
30 categories we have labeled. We therefore also investigate hand contact state,
in particular in both the current frame and in the future (by simply training
on frames before the labeled one). This is an 8-way 
image classification task and we quantify it with accuracy.

\noindent {\bf Models.} We use the same Resnet50 models (pretrained and
fine-tuned). Our labels are sparse temporally; we generate
({\it Pseudo-labels}) by adding the predictions of an initial model on $1M$ training
video frames into the training set. 

\noindent {\bf Results.} We show results in Table \ref{tab:handstate}; for
reference, simply reporting the training mode gets $20\%$. The models
consistently outperform this, even $2$s into the future.
The most common confusions are issues in terms of counting the number of
hands visible. The pseudo-labels consistently give a boost (and training longer
on the same data did not). 

One concern we had was that the system may be exploiting a bias
to solve the contact task. We examined CAM activations
\cite{Zhou2016} and found the network focused on hands and faces; wanting to quantify 
it further, we tried to see if we could decode the features to hand pixel labels.
We freeze the convolutional layers and
learn a linear model on top in a fully convolutional fashion. 
As a proxy to the segmentation, we use the hand bounding boxes.
Using $10$/$100$/$3000$ labeled images for training, this model gets
$31.3$/$40.5$/$47.8$ IoU, substantially outperforming the Imagenet
pretrained model $17.3$/$33.5$/$41.5$. This suggests that the network has indeed 
learned about hands.

\subsection{Hand Detection}
\label{sec:hand_detection}

We then analyze hand detection, testing on other datasets. This
shows that our data is sufficiently varied to work on other datasets
un-finetuned. Additionally, this serves the practical purpose of 
having a detector that works on the wide variety of poses in VLOG, such
as: upper-half views with a 
torso and head (37\%), egocentric-like hands-only views
(31\%), and full body (7\%).

\noindent {\bf Model.} We train a VGG16-based \cite{Simonyan14c}
faster RCNN \cite{Ren2015} using standard settings and joint training.

\noindent {\bf Results.} The most similar dataset with human hands in videos with
both egocentric and third person hands is EgoHands
\cite{Bambach15}, which has a similar number of labeled images (and slightly more
annotated boxes).
EgoHands fails on VLOG, getting $29.5$ AP compared to $67.6$ training
on VLOG. In the other direction, training on VLOG does well, getting
$70.9$ compared to $90.4$ from training on EgoHands (note EgoHands 
tests on people seen at training). As further evidence, 
on a third dataset \cite{Mittal11}, training on VLOG far surpasses using 
EgoHands ($56.3$ vs $31.4$). 

\section{Exploring A Large Set of Hands in Contact}
\label{sec:exploring}

Independent of particular tasks and benchmarks, VLOG represents
a world of humans interacting with objects that
has been indexed in a number of ways. This has obvious
applications in tasks like future prediction \cite{Walker2016}, intuitive
physics \cite{Wu2016}, imitation learning from videos \cite{Yang14}, grasp 
analysis \cite{Huang2015}. We look forward to seeing what can be done with the
dataset, but conclude with a concrete demonstration of the sorts of things
that can be done with a large collection of hands in action: we predict future
locations of hands. 

We build a model that takes an image and predicts the hand locations
$\delta$ frames in the future. This problem has been
tackled in lab settings such as in \cite{Koppula2013}; here,
we do it on large-scale web data. We modify a standard dilated Resnet-54
\cite{Yu17} (details in supplemental) as follows: we introduce a 2-layer network that maps $\delta$
to feature maps; the base feature map and $\delta$ feature map are then concatenated and fused
by 3 convolutional layers to predict hand segmentation. As training data, we run the segmentation model
from Section \ref{sec:hand_detection} on training frames; we trim these to 
$156K$ frames where there is significant change. We then learn
a model for $\delta = 2,\ldots,32$.
Note training this way requires video data from a stationary camera.

We show some predictions in Fig.~\ref{fig:predictions} on held-out data while
varying the timescale $\delta$. Our model has learned reasonable dynamics of where hands go:
in many cases, the hand will continue moving in the direction of likely motion; in others,
such as when humans are holding cell phones, our model has learned that people rarely
put their phones down.

\section{Discussion}

We conclude with a few lessons learnt.  The most
important to us is that explicitly asking for what you want is counterproductive: 
it is easier, not harder, to be implicit. 
In the process, we find a slice of the visual world that is not documented well
by existing explicitly gathered data, including popular datasets. There is good news
though: there are ways to get this data easily and computer vision has reached
a maturity where it can automate much of the process. Additionally, it can
make headway on the difficult problem of understanding hands in action. There is still
a long way to go until we understand everyday interactions, but we believe 
this data puts us on the right path.

\noindent {\bf Acknowledgments.} This work was supported in part by Intel/NSF
VEC award IIS-1539099. We gratefully acknowledge NVIDIA corporation for the
donation of GPUs used for this research. DF thanks David Forsyth, Saurabh
Gupta, Phillip Isola, Andrew Owens, Abhinav Shrivastava, and Shubham Tulsiani
for advice throughout the process.

{\small
\bibliographystyle{ieee}
\bibliography{local}

\begin{thebibliography}{10}\itemsep=-1pt

\bibitem{Alayrac2016}
J.-B. Alayrac, P.~Bojanowski, N.~Agrawal, I.~Laptev, J.~Sivic, and
  S.~Lacoste-Julien.
\newblock Unsupervised learning from narrated instruction videos.
\newblock In {\em CVPR}, 2016.

\bibitem{Bambach15}
S.~Bambach, S.~Lee, D.~Crandall, and C.~Yu.
\newblock Lending a hand: Detecting hands and recognizing activities in complex
  egocentric interactions.
\newblock In {\em ICCV}, 2015.

\bibitem{Burnham78}
K.~P. Burnham and W.~S. Overton.
\newblock Estimation of the size of a closed population when capture
  probabilities vary among animals.
\newblock {\em Biometrika}, 65(3):625--633, 1978.

\bibitem{Carreira17}
J.~Carreira and A.~Zisserman.
\newblock Quo vadis, action recognition? a new model and the kinetics dataset.
\newblock In {\em CVPR}, 2017.

\bibitem{Chao15}
Y.-W. Chao, Z.~Wang, Y.~He, J.~Wang, and J.~Deng.
\newblock Hico: A benchmark for recognizing human-object interactions in
  images.
\newblock In {\em ICCV}, 2015.

\bibitem{Delaitre12}
V.~Delaitre, D.~Fouhey, I.~Laptev, J.~Sivic, A.~Efros, and A.~Gupta.
\newblock Scene semantics from long-term observation of people.
\newblock In {\em ECCV}, 2012.

\bibitem{Heilbron2015}
B.~G. Fabian Caba~Heilbron, Victor~Escorcia and J.~C. Niebles.
\newblock Activitynet: A large-scale video benchmark for human activity
  understanding.
\newblock In {\em CVPR}, 2015.

\bibitem{Fathi2012}
A.~Fathi, Y.~Li, and J.~M. Rehg.
\newblock Learning to recognize daily actions using gaze.
\newblock In {\em ECCV}, 2012.

\bibitem{Fouhey12}
D.~F. Fouhey, V.~Delaitre, A.~Gupta, A.~A. Efros, I.~Laptev, and J.~Sivic.
\newblock People watching: Human actions as a cue for single-view geometry.
\newblock In {\em ECCV}, 2012.

\bibitem{Gibson79}
J.~Gibson.
\newblock {\em The ecological approach to visual perception}.
\newblock Boston: Houghton Mifflin, 1979.

\bibitem{Goyal17}
R.~Goyal, S.~E. Kahou, V.~Michalski, J.~Materzynska, S.~Westphal, H.~Kim,
  V.~Haenel, I.~Fruend, P.~Yianilos, M.~Mueller-Freitag, F.~Hoppe, C.~Thurau,
  I.~Bax, and R.~Memisevic.
\newblock The "something something" video database for learning and evaluating
  visual common sense.
\newblock In {\em ICCV}, 2017.

\bibitem{Grabner11}
H.~Grabner, J.~Gall, and L.~van Gool.
\newblock What makes a chair a chair?
\newblock In {\em CVPR}, 2011.

\bibitem{Gu2017}
C.~Gu, C.~Sun, S.~Vijayanarasimhan, C.~Pantofaru, D.~A. Ross, G.~Toderici,
  Y.~Li, S.~Ricco, R.~Sukthankar, C.~Schmid, and J.~Malik.
\newblock {AVA:} {A} video dataset of spatio-temporally localized atomic visual
  actions.
\newblock {\em CoRR}, abs/1705.08421, 2017.

\bibitem{Gupta11}
A.~Gupta, S.~Satkin, A.~Efros, and M.~Hebert.
\newblock From {3D} scene geometry to human workspace.
\newblock In {\em CVPR}, 2011.

\bibitem{Gupta15}
S.~Gupta and J.~Malik.
\newblock Visual semantic role labeling.
\newblock {\em arXiv preprint arXiv:1505.04474}, 2015.

\bibitem{Hall66}
E.~T. Hall.
\newblock {\em The Hidden Dimension}.
\newblock 1966.

\bibitem{He2015}
K.~He, X.~Zhang, S.~Ren, and J.~Sun.
\newblock Deep residual learning for image recognition.
\newblock In {\em CVPR}, 2016.

\bibitem{Huang2015}
D.-A. Huang, W.-C. Ma, M.~Ma, and K.~M. Kitani.
\newblock How do we use our hands? discovering a diverse set of common grasps.
\newblock In {\em CVPR}, 2015.

\bibitem{huang17}
G.~Huang, Z.~Liu, L.~van~der Maaten, and K.~Q. Weinberger.
\newblock Densely connected convolutional networks.
\newblock In {\em CVPR}, 2017.

\bibitem{JiangSaxena13}
Y.~Jiang and A.~Saxena.
\newblock Hallucinated humans as the hidden context for labeling {3D} scenes.
\newblock In {\em CVPR}, 2013.

\bibitem{Kay17}
W.~Kay, J.~Carreira, K.~Simonyan, B.~Zhang, C.~Hillier, S.~Vijayanarasimhan,
  F.~Viola, T.~Green, T.~Back, P.~Natsev, M.~Suleyman, and A.~Zisserman.
\newblock The kinetics human action video dataset.
\newblock {\em CoRR}, abs/1705.06950, 2017.

\bibitem{Kiberd2015}
R.~Kiberd.
\newblock Youtube's `my daily routine' is a beautiful lie.
\newblock
  http://kernelmag.dailydot.com/issue-sections/staff-editorials/14643/youtube-daily-routine-stepford-wives/.

\bibitem{Koppula2013}
H.~S. Koppula, R.~Gupta, and A.~Saxena.
\newblock Learning human activities and object affordances from {RGB-D} videos.
\newblock {\em The International Journal of Robotics Research}, 32(8):951--970,
  2013.

\bibitem{Krizhevsky12}
A.~Krizhevsky, I.~Sutskever, and G.~E. Hinton.
\newblock Imagenet classification with deep convolutional neural networks.
\newblock In {\em NIPS}, 2012.

\bibitem{Lin2014}
T.-Y. Lin, M.~Maire, S.~Belongie, J.~Hays, P.~Perona, D.~Ramanan, P.~Dollár,
  and C.~L. Zitnick.
\newblock Microsoft coco: Common objects in context.
\newblock In {\em ECCV}, 2014.

\bibitem{Lowe2004}
D.~Lowe.
\newblock {Distinctive Image Features from Scale-Invariant Keypoints}.
\newblock {\em IJCV}, 60(2):91--110, 2004.

\bibitem{Mittal11}
A.~Mittal, A.~Zisserman, and P.~H.~S. Torr.
\newblock Hand detection using multiple proposals.
\newblock In {\em BMVC}, 2011.

\bibitem{Pirsiavash12}
H.~Pirsiavash and D.~Ramanan.
\newblock Detecting activities of daily living in first-person camera views.
\newblock In {\em CVPR}, 2012.

\bibitem{Ren2015}
S.~Ren, K.~He, R.~Girshick, and J.~Sun.
\newblock Faster {R-CNN}: Towards real-time object detection with region
  proposal networks.
\newblock In {\em {NIPS}}, 2015.

\bibitem{Rhinehart2016}
N.~Rhinehart and K.~M. Kitani.
\newblock Learning action maps of large environments via first-person vision.
\newblock In {\em CVPR}, 2016.

\bibitem{Rohrbach2012}
M.~Rohrbach, S.~Amin, M.~Andriluka, and B.~Schiele.
\newblock A database for fine grained activity detection of cooking activities.
\newblock In {\em CVPR}, 2012.

\bibitem{Roy2016}
A.~Roy and S.~Todorovic.
\newblock A multi-scale {CNN} for affordance segmentation in {RGB} images.
\newblock In {\em ECCV}, 2016.

\bibitem{ILSVRC15}
O.~Russakovsky, J.~Deng, H.~Su, J.~Krause, S.~Satheesh, S.~Ma, Z.~Huang,
  A.~Karpathy, A.~Khosla, M.~Bernstein, A.~C. Berg, and L.~Fei-Fei.
\newblock {ImageNet Large Scale Visual Recognition Challenge}.
\newblock {\em IJCV}, pages 1--42, April 2015.

\bibitem{sigurdsson2016much}
G.~A. Sigurdsson, O.~Russakovsky, A.~Farhadi, I.~Laptev, and A.~Gupta.
\newblock Much ado about time: Exhaustive annotation of temporal data.
\newblock In {\em HCOMP}, 2016.

\bibitem{Sigurdsson2016}
G.~A. Sigurdsson, G.~Varol, X.~Wang, A.~Farhadi, I.~Laptev, and A.~Gupta.
\newblock Hollywood in homes: Crowdsourcing data collection for activity
  understanding.
\newblock In {\em ECCV}, 2016.

\bibitem{Simonyan14c}
K.~Simonyan and A.~Zisserman.
\newblock Very deep convolutional networks for large-scale image recognition.
\newblock {\em CoRR}, abs/1409.1556, 2014.

\bibitem{Singh16}
K.~K. Singh, K.~Fatahalian, and A.~A. Efros.
\newblock Krishnacam: Using a longitudinal, single-person, egocentric dataset
  for scene understanding tasks.
\newblock In {\em WACV}, 2016.

\bibitem{UCF101}
K.~Soomro, A.~Zamir, and M.~Shah.
\newblock {UCF101}: A dataset of 101 human actions classes from videos in the
  wild.
\newblock In {\em arXiv preprint arXiv:1212.0402}, 2012.

\bibitem{Sung2012}
J.~Sung, C.~Ponce, B.~Selman, and A.~Saxena.
\newblock Unstructured human activity detection from rgbd images.
\newblock In {\em ICRA}, 2012.

\bibitem{Torralba2011}
A.~Torralba and A.~A. Efros.
\newblock Unbiased look at dataset bias.
\newblock In {\em CVPR}, 2011.

\bibitem{Walker2016}
J.~Walker, C.~Doersch, A.~Gupta, and M.~Hebert.
\newblock An uncertain future: Forecasting from static images using variational
  autoencoders.
\newblock In {\em ECCV}, 2016.

\bibitem{Wang2016}
X.~Wang, A.~Farhadi, and A.~Gupta.
\newblock Actions {\textasciitilde} transformations.
\newblock In {\em CVPR}, 2016.

\bibitem{Wu15}
C.~Wu, J.~Zhang, S.~Savarese, and A.~Saxena.
\newblock Watch-n-patch: Unsupervised understanding of actions and relations.
\newblock In {\em CVPR}, 2015.

\bibitem{Wu2016}
J.~Wu, J.~J. Lim, H.~Zhang, J.~B. Tenenbaum, and W.~T. Freeman.
\newblock Physics 101: Learning physical object properties from unlabeled
  videos.
\newblock In {\em BMVC}, 2016.

\bibitem{Yang14}
Y.~Yang, Y.~Li, C.~Ferm\"uller, and Y.~Aloimonos.
\newblock Robot learning manipulation action plans by “watching”
  unconstrained videos from the world wide web.
\newblock In {\em AAAI}, 2014.

\bibitem{Yao2010}
B.~Yao and L.~Fei-Fei.
\newblock Grouplet: A structured image representation for recognizing human and
  object interactions.
\newblock In {\em CVPR}, 2010.

\bibitem{Yao10a}
B.~Yao and L.~Fei-Fei.
\newblock Modeling mutual context of object and human pose in human-object
  interaction activities.
\newblock In {\em CVPR}, 2010.

\bibitem{Yu17}
F.~Yu, V.~Koltun, and T.~Funkhouser.
\newblock Dilated residual networks.
\newblock In {\em CVPR}, 2017.

\bibitem{Zhou2016}
B.~Zhou, A.~Khosla, A.~Lapedriza, A.~Oliva, and A.~Torralba.
\newblock Learning deep features for discriminative localization.
\newblock In {\em CVPR}, 2016.

\bibitem{zhou17}
B.~Zhou, A.~Lapedriza, A.~Khosla, A.~Oliva, and A.~Torralba.
\newblock Places: A 10 million image database for scene recognition.
\newblock {\em TPAMI}, 2017.

\end{thebibliography}
}

\appendix

\section{Acquisition Pipeline Details}

This is somewhat involved and difficult, but primarily an engineering, not research task. It is documented
here to answer technical questions.

\noindent {\bf Finding Videos.} YouTube provides, as an undocumented feature, four thumbnails.
We download these and extract Alexnet {\tt pool5} features on these thumbnails. 
Our feature vector is the average {\tt pool5} feature plus the min, mean, and max distance between 
the activations. These distance features help find videos where someone is
talking to the camera the entire time.

\noindent {\bf Finding Episodes within Videos.} We use an approach based on SIFT homography fitting
after experimenting with a number of alternate approaches. Our method scans every 10 frames, looking
for unmatchable frames and large movements. Since we are especially interested in static clips and many
clips are already static, we first look for support for the identity
transformation, and only then start RANSAC iterations. This saves considerable computation.

\begin{enumerate}
\item We first run shot detection every 10 frames ($\sim 3$Hz)
\item We then scan forward every 30, 60, and 90 frames (where the clip is long enough for this), and verify that there
is some evidence of a match. Frames that cannot be matched forward tend to be during dissolves, which are frequent enough
to require removal. We mark a discontinuity at any point where the frame cannot be matched.
\item Finally, given the shot boundaries at every 10 frames, we examine the 10 frames in the middle. This fine-scale
detection is crucial: many videos are aggressively trimmed and not doing this results in premature cuts.
\end{enumerate}

For $N$ frames that are segmented into $k$ shots, this procedure requires
$N/10 + 10*(k-1)$ SIFT feature extractions as opposed to $N$,
and similar savings on homography fits.

\noindent {\bf Finding Clips.} We re-run the video classifier; here, since the
much shorter clips typically have very similar interframe appearance, we retrain the
classifier using only the average CNN activation feature, and not the
inter-frame distance features.

\begin{figure*}
\centering
\includegraphics[width=\linewidth]{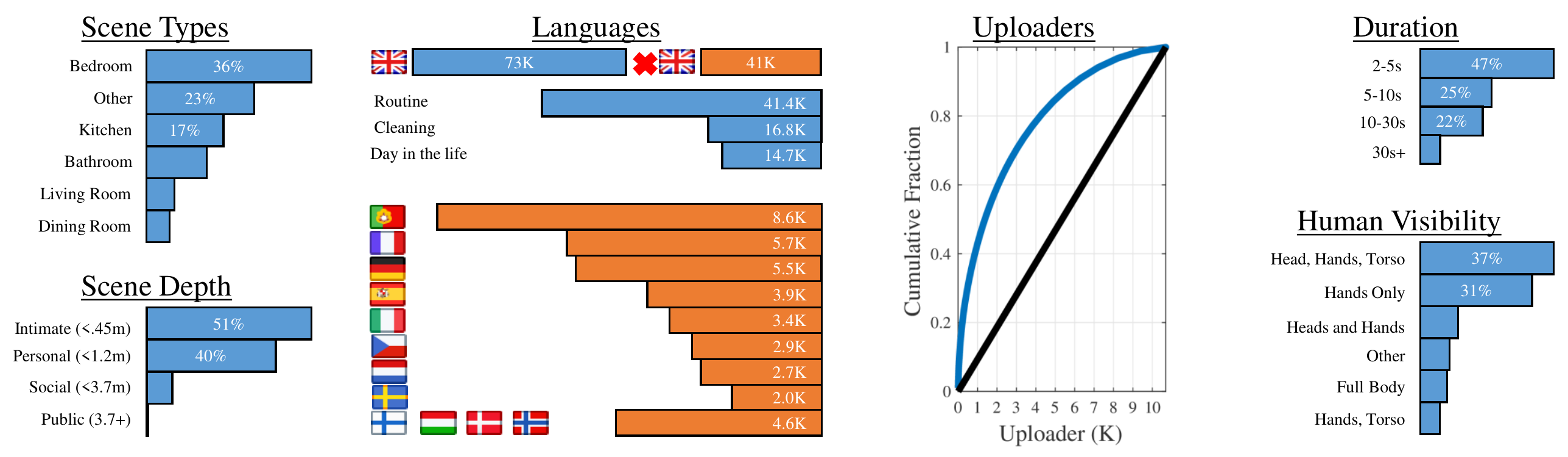}
\caption{Additional statistics about VLOG}
\label{fig:appendix_stat}
\end{figure*}

\section{Additional Experimental Details, Statistics}

We now provide some supplemental details on a few of our
experiments.

\subsection{Additional statistics}

Figure \ref{fig:appendix_stat} shows  additional statistics about VLOG: where it comes from, the distribution of types (scene depth and class), a CDF 
of uploaders, video lengths (average length $\approx 10$s), and the visibility of various body types.

\subsection{Entry-level categories in our sample}

We provide the name (post canonicalization) of the objects that humans touch and their count in our
500 image sampling of the dataset. We note that many of the objects 
are probably more naturally referred to more specifically (e.g, ``shampoo'' instead
of ``bottle'', ``crackers'' instead of ``box''). Note the large number of common, natural
categories that simply appear once: plant, zucchini, bathtub, etc. Note also that this does not include any verbs.

\noindent {\bf Categories:} bottle (36), cellphone (31), makeupbrush (16), blanket (14), laptop (14), spoon (13), bag (12), body (12), jar (11), notebook (10), tube (10), book (9), box (9), makeupcompact (9), shirt (9), bowl (8), knife (8), paper (8), table (8), toothbrush (7), pillow (6), cabinet (5), cup (5), drawer (5), fork (5), glas (5), hairstraightener (5), lipstick (5), pan (5), shoe (5), towel (5), backpack (4), bread (4), doll (4), door (4), facewipe (4), hairbrush (4), mug (4), banana (3), bed (3), carton (3), meat (3), mop (3), nailpolish (3), plate (3), refrigerator (3), sheet (3), sink (3), sponge (3), sweatshirt (3), toothpaste (3), cat (2), chair (2), cheese (2), counter (2), dish (2), dog (2), facebrush (2), floor (2), hairdryer (2), hanger (2), iron (2), monitor (2), pen (2), pitcher (2), pot (2), remote (2), spraybottle (2), stuffedanimal (2), toy (2), yogamat (2), apple (1), armwarmer (1), baby (1), babyjumper (1), basket (1), bathtub (1), blender (1), bookshelf (1), butter (1), cage (1), calculator (1), car (1), cheesegrater (1), coffeetable (1), container (1), cookie (1), diaper (1), dirt (1), dishwasher (1), dresser (1), drill (1), drumstick (1), dumbbell (1), duster (1), egg (1), espressomachine (1), facialmask (1), flower (1), folder (1), gamecontroller (1), icecubetray (1), jewelryholder (1), kettle (1), lamp (1), lettuce (1), lid (1), magazine (1), measuringcup (1), muffin (1), napkin (1), nightstand (1), pacifier (1), paintbrush (1), paintroller (1), pencil (1), piano (1), pie (1), plant (1), popsiclemold (1), powercord (1), purse (1), rag (1), sand (1), sander (1), sewingmachine (1), shelf (1), skirt (1), stove (1), straw (1), tape (1), tin (1), tray (1), wardrobe (1), watch (1), zucchini (1).

\subsection{Hand prediction} 

Here we provide the architecture of the hand prediction network. Let: $C(k,s)$ denote a convolution
of $k$ kernels with size $s \times s$; $R$ denote ReLU and $BN$ denote Batchnorm.
\begin{enumerate}
\item The image $I$ is passed through the base DRN-D-54 network $\phi$, yielding a $512$ channel
feature map. 
\item The time offset variable $\delta$ is upsampled to feature map size, then mapped to $64D$ through two layers ($C(16,1) \to R \to C(64,1) \to R$),
or in total $\psi$.
\item After concatenating image $\phi(I)$ and time features $\psi(\delta)$, we predict the final output passing it through three
$3 \times 3$ convolutions (denoted $\zeta$ in total): $C(128,3) \to BN \to R \to C(128,3) \to BN \to R \to C(2)$, followed by 
$8\times$ bilinear upsampling. 
\end{enumerate} 
In total the network is $\zeta(\textrm{cat}(\phi(I),\psi(\delta)))$. The network is trained to minimize a cross-entropy loss.

\end{document}